\documentclass[letterpaper, 11 pt, conference]{ieeeconf}      
\IEEEoverridecommandlockouts 

\usepackage[utf8]{inputenc} 
\usepackage[hidelinks]{hyperref}       
\usepackage{amsmath}
\usepackage{amsfonts} 
\usepackage{statmath}
\usepackage{authblk}

\title{\LARGE \bf Interpretability  of Neural Network With Physiological Mechanism}
\author[1*]{Anna Zou\thanks{*Corresponding author: Anna Zou. Email: zoua@ufl.edu}\thanks{This research received no external funding by any organization}}
\author[2,3]{Zhiyuan Li}
\affil[1]{Department of Psychology, University of Florida}
\affil[2]{Department of Electrical Engineering and Computer Science, University of Cincinnati}
\affil[3]{Imaging Research Center, Cincinnati Children's Hospital Medical Center}

\begin{document}
\maketitle

\begin{abstract}
Deep learning continues to be a powerful state-of-art technique that has achieved extraordinary accuracy levels in various domains of regression and classification tasks, including image, signal, and natural language data. The original goal of proposing the neural network model is to improve the understanding of complex human brains using a mathematical approach. However, recent deep learning techniques continue to be difficult to interpret in addition to challenges in explaining its functional process. As a result it is being treated mostly as a black-box approximator. Deep learning techniques have continued to steer further and further away from the realistic human brain model, despite its original intents. Such an AI model needs to be biologically and physiologically realistic in order to incorporate a greater understanding of human-machine evolutionary intelligence. In this paper, we compare neural networks with biological mechanisms and physiology to discover the similarities and differences between ANN and what we currently know about the human brain. We also attempt to explain ANNs by exploring human biological behaviors and brain anatomy. 
\end{abstract}


\section{Introduction}
Artificial neural networks (ANNs), which is inspired by human biological brain process, has been widely utilized in areas of deep learning research, including regression, classification, pattern recognition, clustering and prediction in many disciplines. The scale of ANNs has became more and more competitive to predict the outcomes of interests, such as multilayer perceptrons (MLPs) \cite{rosenblatt1961principles}, convolutions neural networks (CNNs) \cite{lecun2015deep}, and long short-term memory (LSTM) \cite{hochreiter1997long}. These derivatives of ANNs are proposed using mathematical expression to learn the diverse representations of data type, including cross-sectional and spatiotemporal. However, the interpretation of deep neural networks (DNNs) solutions remains poorly understood, resulting many to treat these data models as a black box, and to doubt whether DNNs can be explained at all \cite{lipton2018mythos,shwartz2017opening,koh2017understanding,mishra2014view}.

The original goal of proposing ANNs was to seek for potential mathematical explanations to improve our understanding of complex brain functions \cite{rosenblatt1958perceptron}. To address these goals, such AI models need to be biologically realistic. Despite recent advances in the development of ANNs and their ability to perform complex cognitive and perceptual tasks like humans, their similarity to aspects of brain anatomy is imperfect. Some imperfections in relation to biological mechanisms, i.e., performing the backpropagation (BP) \cite{rumelhart1985learning}, which is a computation algorithms for an input-output instance, including gradient descent, to efficiently minimize the errors of loss with respect to the weights or coefficients parameters of the networks. BP practices fine-tuning of a neural network based on the error rate (loss) at the previous iteration (epoch). This process works in a backwards path, which is often not depicted in the activities of real human brain neurons. Within human brains, information is transmitted by action potentials via neurons and through synapses. Action potential is a series of electrical events the nerve cells use to communicate information from neuron to neuron \cite{barnett2007action}, it only move in one direction, from the cell body to the presynaptic terminals. During the refractory period, inactivated sodium channels in previous axon segments prevent the membrane from depolarizing as the action potential moves from one Node of Ranvier to the next \cite{dodge1973action}. As a result, the action potential can only propagate toward axon segments with closed sodium channels \cite{barnett2007action,bean2007action,dodge1973action}. Additionally, connectivity profiles and gene expressions are intricate and often differ with each type of neuron \cite{elston2003cortex,tasic2018shared}. Cortical regions receive and send signals to and from many cortical and sub-cortical regions \cite{lillicrap2020backpropagation}, all of which are difficult and often not interpreted or simplified on ANN model.

In deep learning, loss function represents the objective function that evaluates the learning performances from the empirical instances, thereby, the learning algorithms are seeking to minimize this objectives function by searching for the optimal solution of weight parameters for the model \cite{wang2020comprehensive,sugiyama2012machine}. In statistics and decision theory, loss function or risk function can be treated as parameter estimation that is to obtain the difference between the estimated value and the ground-truth value \cite{nematollahi2009estimation,silverman2018density}. Marblestone, A. H., et al. \cite{marblestone2016toward} suggests one aspect of deep learning models that can be explained and observed in the brain's physiology which is the loss function. Specifically, the ability of human brain to optimize the loss functions. The general idea of \cite{marblestone2016toward} argued that the loss function can be treated as learning amongst human species. Minimizing the risk to achieve a higher probability of making the correct or most optimal behaviors in a domain. Human brain and human behaviors often simulate various loss functions within the realm of deep learning. Such that, humans are unconsciously trying to minimize their loss function in day-to-day activities including efforts to minimize personal risks. This suggest that our brains learn the most favorable strategies to employ in a given situation to avoid adverse events amongst many things. In this paper, we aim to address the following question: what are some similarities and differences between ANN models and the human brain and behaviors?

\section{Physiological view of neural networks}
A standard framework of neural network models contain three main components: an input layer, a hidden layer and an output layer, and each layer can be interpreted from different perspectives. In this section, we focus on the explanation of neural networks not only in mathematical expression but incorporating biological perspectives.

\subsection{Formulation of Neural Networks}
Consider a supervised learning task using a single layer unit neural network: suppose we have a feature matrix $\bm{X}=\{\bm{x}_i\}_{i=1}^{N}$ with label $\bm{y}=\{y_i\}_{i=1}^{N}$, where N is the number of training instances. To predict $\bm{y}$ using $\bm{X}$, the algebraic representation of basis function is defined as 
\begin{align}
    f_{\bm{w},\bm{b}}(x)&=\sigma(\sum_{i=1}^{N}\bm{w}_i\bm{x}_i+\bm{b})
\end{align}
where $\sigma(\cdot)$ is the activation function, $\bm{x},\bm{w}$ and $\bm{b}$ represents the input, weight and bias, respectively. From biological or neuroscience perspective, each neuron consist of dendrites, soma, and axon, in which dendrites, receives information from other neurons, transmit electrical stimulation to the soma, and the information is transmitted to outside by axon. This typical biological process illustrated how human brain networks performs, thereby, the mathematical expression of equation (1) can be interpreted by the function of human brain neurons if we represent the dendrites as input layer, the soma as hidden layer and the axon as output layer. 

\subsection{Loss Function and Human Behavior}
Performing a highly precise prediction is equivalent to minimizing the error of loss through searching for optimal decisions in human behavior \cite{marblestone2016toward}. The only criteria by which choices should be selected equals to how far our expectation is from the real goal. Loss functions measures the divergence or the distance of probabilistic distribution between predicted values and its true values. Loss functions are not fixed, depends on the interest of tasks, it usually divides into two main tasks: regression and classification. In here, we interpret loss functions using the perspective of human behaviors. 

\subsubsection{Regression Loss}
Suppose $\bm{y}$ follows a continuous probability distribution, i.e., Gaussian, the goal of equation (1) becomes to estimate the specific value, denoted as $\hat{\bm{y}}$, which is continuous in nature. In regression task, we usually take distance measures as the loss function, such that 
\begin{align}
    \text{MAE}&=\frac{\sum_{i=1}^{N}|\bm{y_i}-\hat{\bm{y}}|}{N} \\
    \text{RMSE}&=\sqrt{\frac{\sum_{i=1}^{N}(\bm{y_i}-\hat{\bm{y}})^2}{N}}
\end{align}
where MAE and RMSE represents the mean absolute error ($L_1$ loss) and root mean squared error ($L_2$ loss), individually. These two loss functions both measure the difference between the predictions and the ground-truth. 

\subsubsection{Classification Loss}
Classification task involve classify the output into different groups or clusters where the number of group can be either binary or more than two. For binary classification task, assume $\bm{y}\sim \text{Bernoulli}(p)$, the probability density function of $\bm{y}$ is given by 
\begin{align}
    f(y)&=p ^ {y} (1 - p) ^ {1 - y}, y=0,1
\end{align}
where $p\in (0,1)$ indicates the probability that output belongs to the first group. To model $\bm{y}$ using observed data $\bm{X}$, the log-likelihood function is then defined as
\begin{align}
    l(\bm{p}|\bm{y})&=log\prod_{i=1}^{N}p_i^{y_i}(1-p_i)^{1-y_i}  \\
                         &=\sum_{i=1}^{N}y_ilogp_i + (1-y_i)log(1-p_i) \nonumber
\end{align}
where $p_i$ equivalent to $f_{\bm{w},\bm{b}}(x)=\sigma(\sum_{i=1}^{N}\bm{w}_i\bm{x}_i+\bm{b})$ in equation (1), in practice, $\sigma(z)=\frac{1}{1+exp(-z)}$ is the sigmoid function in order to learn the non-linear decision boundary for separating two data groups. Since $l(\bm{p}|\bm{y})$ is the log-likelihood function, we seek to maximize $l(\bm{p}|\bm{y})$ for estimating the optimal parameter $\bm{w}_i$, this problem is formulated into minimizing the $-l(\bm{p}|\bm{y})$, which is equivalent to 
\begin{align}
    \mathcal{L}=-\frac{1}{N}\sum_{i=1}^{N}y_ilogp_i + (1-y_i)log(1-p_i) 
\end{align}
where $\mathcal{L}$, in here, we call it as cross-entropy loss. Without loss of generality, for multi-class classification, we minimize $\mathcal{L}$ in a more general expression, which is given by
\begin{align}
    \mathcal{L}^{*} = -\frac{1}{N}\sum_{i=1}^{N}y_ilogp_i
\end{align}
where the activation function in here will use the softmax function, i.e., $\sigma(z)=\frac{exp(z_i)}{\sum_{j=1}^{N}exp(z_j)}$ of $f_{\bm{w},\bm{b}}(x)$.

\subsubsection{Interpretation}
According to Marblestone, A, H., et al. \cite{marblestone2016toward}, if applying the idea of loss functions to the human brain, such idea can be explained by neuroplasticity or brain plasticity where neurons in a brain area can change their properties. For example, synaptic properties. The Hebbian theory states that an increase in synaptic efficacy is attributed to repeated and persistent stimulation of postsynaptic cells by presynaptic cells \cite{hebb2005organization}. The purpose is to explain synaptic plasticity, or what happens to neurons during learning. Synaptic plasticity occurs at the single-cell level and is a form of neuroplasticity. Marblestone, A, H., et al. \cite{marblestone2016toward} suggests that optimized strategies may have been learned by the brain which is similar to the deep learning loss function where human brains change and adapt throughout life in response to experiences. Furthermore, there are countless different types of loss function within deep learning, some are much closer to the realm of biological mechanisms than we anticipate. For example, classification loss involves predicting a discrete class output. Such classifications can be seen in human decision-making. Specifically, how humans come to decisions, make judgments, solve problems, etc. Heuristics is often employed by humans to solve trivial problems encountered and further allows humans to make informed judgement calls, and it can be heavily altered by current emotions, available information, and more \cite{gigerenzer2008heuristics,sunstein2005moral,michalewicz2013solve}. However, a difference between machine and human processes exists that heuristics does not improve with continued experience, exposure, or training, unlike loss classification. 

\subsection{Learning Rules}
Searching the weight parameters, i.e., $\bm{w}_i$, is the essential goal to develop a neural network with high generalizability. To learn $\bm{w}_i$ of the neural network, we aim to minimize the cross entropy loss $\mathcal{L}^{*}$ that represents a function of $\bm{w}_i$. In deep learning area, BP is the widely used numerical algorithm for learning $\bm{w}_i$ of the neural network. Here, we review two learning theories: Hebbian learning rule and iterative optimization. Then, we discuss the difference between the physiology of neurons and BP.

\subsubsection{Hebbian Learning}
Biological neural weight adjustment is the basis of Hebbian Learning. The purpose of Hebbian learning rule aims to quantify the proportion difference of the weight of the connection between two neurons, and describes that how neuron learns and develops the cognition with response to the outside stimuli \cite{miller1994role, gerstner2002mathematical,kempter1999hebbian}. The formulation of Hebbian learning rule is defined as 
\begin{align}
    \bm{w}_i &\leftarrow \bm{w}_i + \bm{x}_iy_i \\
     b_i &\leftarrow  b_i + y_i
\end{align}
where the $\bm{w}_i$ is the weight connection between two neurons, and it continues to be updated between neurons in the neural network for each training instances. This algorithm basically states that the strength of synapse will be increased if the connection between two neurons are activated. 

\subsubsection{Learning as Optimization}
Comparing to Hebbian learning rule, the iterative optimization algorithms, including gradient descent, are very similar with the only difference is numerical optimization usually consider the gradient of learning parameters. Consider a deep neural network with $K$ layers with a given an input matrix $\bm{X}$, the feed-forward output is defined as 
\begin{align}
    \hat{\bm{y}} &= f^{K}(\bm{W}^K f^{K-1} (\bm{W}^{K-1}\dots f^1(\bm{W}^1\bm{X})))
\end{align}
where $f^k$ and $\bm{W}^{k}$ be the activation function and weight matrix at layer $k$, $k\in K$, respectively. Thereby, the learning objective loss function is then given by
\begin{align}
    \mathcal{L}_{\bm{W}} &= R(\bm{y}, \hat{\bm{y}})
\end{align}
where $R(\cdot)$ is the loss function between true value $\bm{y}$ and estimated value $\hat{\bm{y}}$. To learn $\bm{W}^{k}$, $k\in K$ of each layer, a general iterative algorithm (e.g., gradient descent) is usually applied, such that
\begin{align}
    \bm{W}^{k}\leftarrow \bm{W}^{k}-\eta \nabla \mathcal{L}_{\bm{W}^{k}}
\end{align}
where $\eta$ is the learning rate, and $\nabla \mathcal{L}_{\bm{W}^{k}}=\frac{\partial \mathcal{L}_{\bm{W}^{k}}}{\partial \bm{W^k}}=\frac{\partial \mathcal{L}_{\bm{W}^{k}}}{\partial f^k}\frac{\partial f^k}{\partial \bm{W^k}}$ is the gradient of $\mathcal{L}_{\bm{W}^{k}}$ using BP. This iteration will stop when the error loss is converged into a small number. 

\subsubsection{Backpropagation and Action Potential}
Synaptic connections between neurons are modified as a result of learning. In deep learning, researchers use synaptic updates to improve the performance of neural networks without being restricted by biological reality. For BP error computations, it start in the output layer of the neural networks and moves backwards, leading to the errors ‘backpropagating’ through the network. With relating to neuron physiology there lacks direct evidence that the brain uses such BP algorithm or method for learning. As previously mentioned, BP feeds error computations from the output layer, moving backwards. Thus, we may interpret the weight or error computation as information. When looking at relative comparisons in neuron physiology, action potential carries out similar responses where information is carried out through the neuron. However, a difference exists between BP and human neurons. Within human neurons our 'information' or nerve impulses is rarely transmitted backwards like it does in BP. The reason for such is due to the refractory period. The refractory period will result in parts of axons whose recently generated an action potential to be unresponsive \cite{farmer1960refractory}. In other words, the traveling action potential is unable to generate another action potential in the retrograde direction, because the only available region for excitement is in the anterograde direction to the terminal. When an action potential has reached the end of the axon terminal, this stimulates a release of neurotransmitters from the presynaptic terminals, across the synapse, to receptor site receiving the information \cite{flores2012trafficking, todd2010gap,pereda2013gap}.
Under a unique circumstance, gap junction between two neurons are essentially electrically linked together, making BP theoretically possible in the brain. However, it is also important to question whether the term BP is even applicable to such situation. The presynaptic terminal contains and releases neurotransmitters that bind onto receptors found on postsynaptic neurons. BP also seems improbable in this scenario because neurotransmitters are located at the synaptic vesicles of the presynaptic neuron and the receiving receptors for the released neurotransmitters are located in the postsynaptic \cite{fukuda2007structural}. Presynaptic neurons also contain neurotransmitter receptors, but with different purpose of those present on postsynaptic neurons such that presynaptic neurotransmitters are used to prevent further release of a neurotransmitter.

\section{Discussion}
Neural networks differ from AI models in that its structure is inspired by neurons within the human brain. In this paper, we analyzed the mathematical expression of neural networks and interpreted the loss functions employed in both computational models and human behaviors. We further analyzed the idea behind aspects of neural network such as the loss functions that can be understood in humans when looking for the optimal decision in a situation. Principles of loss function can be observed by synaptic plasticity such as in the Hebbian theory where strengthening of the synapses increases efficiency due to constant stimulation of presynaptic and postsynaptic cells. Similar to deep learning, loss function is continuously updating and 'learning' to minimize its loss, in humans our brains are repeatedly adapting, changing and learning optimal strategies in response to exposure to experiences. If looking at classification loss specifically, which involves distinct class groupings or clusters, such operation can be translated to how humans make decisions, judgements and solve problems. An example from psychology include the ideas of heuristics which are mental shortcuts allowing people to efficiently make judgement calls and solve trivial problems \cite{gigerenzer2011heuristic,gilovich2002heuristics}. However, such heuristic differs from loss classification because heuristics are not improving with continued training or experience. Furthermore, we also analyzed the learning rules within neural networks starting from the Hebbian learning rule presented by \cite{kempter1999hebbian} to iterative numerical optimization. Specifically, how both learning rule methods initialize error computations in the output layer of the neural network and work towards a backwards direction, often known as BP. BP practices calibrating the weights of neural network based on the loss obtained from previous iteration. In this paper, we sought out how the events of BP is often not observed within neuron physiology. Such that, we can treat the weight and action potential in a neuron as information, respectively. If looking at it this way, nerve impulses rarely transmit retrograde due to the refractory period of a neuron. The refractory period is the amount of time necessary for an excitable membrane to be able to respond to another stimulus. Furthermore, we analyzed the gap junction between neurons in efforts towards a better consideration between BP in neural networks and biological neural circuits. It was observed that though two neurons can technically be electrically joined together through the gap junction \cite{connors2004electrical}, making BP theoretically possible, but the present physiological aspect of the gap junction proves principles of BP improbable. With neurotransmitters being released from the synaptic vesicles of the presynaptic terminal after arrival of nerve impulses, and with neurotransmitter receiving receptors only being present on postsynaptic neurons, it makes the operation of BP unfeasible.
Despite our efforts to present a versatile comparison between physiology and neural network, there still exists many limitations to our knowledge in terms of how well the networks truly constitute for human biological neural circuits. Specifically if certain algorithms or models such as BP are better applicable for specific regions of the brain \cite{zipser1988back,lillicrap2013preference,cadieu2014deep} where such models can account for certain observed neural responses. Future research should continue efforts in improving the similarities between biological neural circuits, the brain and neurocomputational models. Perhaps applying constrains to computational models of individual brains or populations. Computational modeling, after all present us with a wide variety of AI models such as ANN explained in this paper, but often lack natural human constrains such as action potential frequencies and signal travel speeds. Applying constrains to neural networks can potentially allow for new perspectives and novel findings such as a better understanding of our neural circuitry. In addition, applying constrains allow for more biologically realistic models that can provide aid in individualizing therapy plans, surgery, treatments and more. 

\section{Conclusion}
In summary, the aim of this paper is to investigate the interpretability of neural networks in a physiological view and compare the functional process between AI model and human behavior and brain. We discussed the perspective of learning rules in both Hebbian learning and numerical optimization and provide interpretation using the mechanisms of action potential. The gap junction was also analyzed and discussed in comparison to operations of BP. It is important to not only employ deep learning techniques in various domains considering its current successes but it is also important to explore its similarity and differences to the human brain and behaviors. For greater knowledge and achievements in the future to the realms of neuroscience, efforts need to be made to improve the biological realisticness of these models.

\bibliographystyle{unsrt}
\bibliography{myref.bib}
\end{document}